\documentclass[a4paper, 11pt]{article}
\usepackage[margin=25mm]{geometry}

\usepackage{natbib}
\usepackage{apalike}
\usepackage{xpatch}
\usepackage{amsfonts}
\usepackage{amsmath}
\usepackage{amssymb}
\usepackage{amsthm}
\usepackage{bbm}
\usepackage{booktabs}
\usepackage{graphicx}
\usepackage{caption}
\usepackage{enumitem}
\usepackage{hyperref}
\hypersetup{
    colorlinks=true,
    citecolor=blue,
    linkcolor=blue
}

\title{Bayesian Semi-supervised Multi-category Classification under Nonparanormality}
\author{Rui Zhu$^{1}$, Shuvrarghya Ghosh$^{2}$, Subhashis Ghosal$^{2}$  \\
        \small $^{1}$ Google Inc. \\
        \small $^{2}$ North Carolina State University \\
}

\providecommand{\keywords}[1]
{
  \small	
  \textbf{\textit{Keywords---}} #1
}

\DeclareMathOperator*{\argmin}{\arg\,\min}
\DeclareMathOperator*{\argmax}{\arg\,\max}

\newcommand{\N}{\mathrm{\text{N}}}
\newcommand{\TN}{\mathrm{\text{TN}}}
\newcommand{\p}{\mathrm{P}}
\newcommand{\W}{\mathrm{\text{W}}}
\newcommand{\ob}{\mathrm{ob}}
\newcommand{\new}{\mathrm{new}}

\makeatletter
\xpatchcmd\NAT@citex
 {%
  \@citea\NAT@hyper@{%
    \NAT@nmfmt{\NAT@nm}%
    \hyper@natlinkbreak{\NAT@aysep\NAT@spacechar}{\@citeb\@extra@b@citeb}%
    \NAT@date
  }%
 }
 {%
  \@citea
  \NAT@nmfmt{\NAT@nm}%
  \NAT@aysep\NAT@spacechar
  \NAT@hyper@{\NAT@date}%
 }
 {}{}
\xpatchcmd\NAT@citex
 {%
  \@citea\NAT@hyper@{%
    \NAT@nmfmt{\NAT@nm}%
    \hyper@natlinkbreak{\NAT@spacechar\NAT@@open\if*#1*\else#1\NAT@spacechar\fi}%
    {\@citeb\@extra@b@citeb}%
    \NAT@date
  }%
 }
 {
  \@citea
    \NAT@nmfmt{\NAT@nm}%
    \NAT@spacechar\NAT@@open\if*#1*\else#1\NAT@spacechar\fi
    \NAT@hyper@{\NAT@date}%
 }
 {}{}
\makeatother

\begin{document}
\maketitle

\begin{abstract}
Semi-supervised learning is a model training method that uses both labeled and unlabeled data. This paper proposes a fully Bayes semi-supervised learning algorithm that can be applied to any multi-category classification problem. We assume the labels are missing at random when using unlabeled data in a semi-supervised setting. Suppose we have $K$ classes in the data. We assume that the observations follow $K$ multivariate normal distributions depending on their true class labels after some common unknown transformation is applied to each component of the observation vector. The function is expanded in a B-splines series, and a prior is added to the coefficients. We consider a normal prior on the coefficients and constrain the values to meet the normality and identifiability constraints requirement. The precision matrices of the Gaussian distributions are given a conjugate Wishart prior, while the means are given the improper uniform prior. The resulting posterior is still conditionally conjugate, and the Gibbs sampler aided by a data-augmentation technique can thus be adopted. An extensive simulation study compares the proposed method with several other available methods. The proposed method is also applied to real datasets on diagnosing breast cancer and classification of signals. We conclude that the proposed method has a better prediction accuracy in various cases.
\end{abstract}

\section{Introduction}\label{sec:intro}

Semi-supervised learning is a method that uses labeled and unlabeled data to train the classifier. There has been a growing interest in semi-supervised learning in recent years. Traditional classification methods are usually supervised and use only labeled data for training. We need the actual labels for all units in the dataset and the associated measurements to train a traditional classifier. However, labels are often difficult, expensive, or time-consuming to obtain and require a lot of human labor and expertise. For example, an algorithm can quickly alert a fraud detection system, but the actual status verification may not be pursued. One way to identify fraudulent transactions or account takeovers is through client reports. However, it is unlikely for clients to notice every crime act. There could be a considerable amount of undetected fraud if we rely on the labeling process only on client reports. The only way to obtain a thorough and reliable label for every instance is through human evaluation, meaning that a verification team will give a judgment given the information, which will cost both time and human labor. On the other hand, unlabeled data, in most cases, are substantial and easy to obtain. Unsupervised learning methods like clustering provide a way to use unlabeled data. But, it may not be appropriate or helpful in a classification problem. Supervised learning seems to be the best solution for these cases. When a few units in a dataset have labels, the semi-supervised learning technique can use these large numbers of unlabeled units in classification instead of discarding them, thus significantly improving the classifier's performance. Semi-supervised learning lies between supervised and unsupervised learning, and the goal is to make the most efficient use of data containing labeled and unlabeled observations.

Semi-supervised learning can be classified into two general categories \textemdash
 generative methods and discriminative methods. Generative methods are often based on the Expectation-Maximization (EM) algorithm of \citet{Dempster1977} and make assumptions on the underlying distribution for different classes. Discriminative methods, on the other hand, only learn the boundary between classes. In recent literature, discriminative methods have been explored more than generative methods. Self-training (\citet{Yarowsky1995}) is the most widely-used semi-supervised learning method. The basic idea is that the chosen classifier teaches itself the labels and then learns iteratively. Co-training (\citet{Blum1998}) is conducted by splitting the features into two subsets. The idea is that each subset can learn and teach the other some labels they are confident of. Another commonly used algorithm is Semi-Supervised Support Vector Machines ($\mathrm{S^3VM}$). The goal of $\mathrm{S^3VM}$ is to find a linear boundary with the maximum margin for labeled and unlabeled data. This is an NP-hard problem, and many algorithms are proposed to solve it. In our simulations, we have compared our method to several different SVM algorithms like that of \citet{Belkin2006}, \citet{Liu2012}, \citet{Sindhwani2006}. The method $\mathrm{S^3VM}$ is based on the assumption that the boundary will not cut in dense regions. Other methods like the Gaussian Process (\citet{Lawrence2005}) also use this assumption. Methods based on graphical analysis are proposed in \citet{Balcan2005}, \citet{Zemel2005}, \citet{Zhang2007}, \citet{Hein2007}. A thorough introduction to semi-supervised learning methods can be found in \citet{Zhu2005}.

Generative methods based on the EM algorithm have two widely known disadvantages. First, some assumptions on the underlying distributions have to be made. If the assumptions are wrong, the mislabeled data will hurt the accuracy \cite{Cozman2003}. Second, the EM algorithm tends to stick to a local maximum instead of the global maximum. This may also cause problems when using unlabeled data. 

In this paper, we propose a new generative method for semi-supervised learning which can solve the two problems mentioned. First, we make a flexible semi-parametric modeling assumption. We assume the two underlying distributions map to multivariate normal distributions after the same componentwise unknown transformation on each class. This is more general than a specific parametric assumption and is called the nonparanormal assumption (\citet{Liu2012, Mulgrave2020}). \citet{Mulgrave2020} used this to generalize a Gaussian graphical model. Thus, we are in a semiparametric setting since the transformation has an unrestricted functional form. We use a random series expansion based on B-splines to put a prior on the transformation function and then an appropriate prior on the coefficients of B-splines with monotonicity and identifiability constraints. Instead of finding point estimates using the EM algorithm, we obtain the posterior distributions of the unlabeled data using a Gibbs sampling framework. This prevents the problem of trapping at local maxima. 

The rest of the paper is organized as follows. We describe the necessary notations, the model, and the algorithm given by Gibbs Sampling in Section \ref{sec:model}. We give a method of selecting hyperparameters in Section \ref{sec:select}. Then, we present the results of a simulation study to compare our method with other semi-supervised learning methods in Section \ref{sec:binsim} in a binary classification setup. We apply our method to some real binary classification data sets in Section \ref{sec:binreal}. Finally, in section \ref{sec:multsim}, we present a simulation study to compare the proposed method with supervised learning methods in a multi-label classification setup and conclude the paper with Section \ref{sec:diss}.

\section{Model} 
\label{sec:model}

\textbf{Notations}: 
Let $\N_p(\boldsymbol{\mu}, \boldsymbol{\Sigma})$ denote a $p$-dimensional normal distribution with mean $\boldsymbol{\mu}$ and covariance $\boldsymbol{\Sigma}$; $\Phi(\cdot; \boldsymbol{\mu}, \boldsymbol{\Sigma})$ and $\phi(\cdot; \boldsymbol{\mu}, \boldsymbol{\Sigma})$ denote the cumulative distribution function and the probability density function of a normal distribution with mean $\boldsymbol{\mu}$ and covariance $\boldsymbol{\Sigma}$ respectively; $\TN_p(\boldsymbol{\mu}, \boldsymbol{\Sigma}, \mathcal{T})$ stands for a truncated normal distribution $\N_p(\boldsymbol{\mu}, \boldsymbol{\Sigma})$ with its support restricted on a set $\mathcal{T}$; $\W_p(n, V)$ stands for the Wishart distribution with $n$ degree of freedom, $n > p-1$, and a positive definite scale matrix $V$ of order $p\times p$; $\mathrm{Beta}(\alpha, \beta)$ stands for the beta distribution with shape parameters $\alpha$ and $\beta$.

Suppose we observe $\boldsymbol{X}^{(1)}, \ldots, \boldsymbol{X}^{(n)}$ independently, which take value in $\mathbb{R}^p$ for some $p \geq 1$, each observation as $\boldsymbol{X}^{(i)}=(X_1^{(i)}, \ldots, X_p^{(i)})$. 
We denote the class for $\boldsymbol{X}^{(i)}$ by $L^{(i)}\in \{1, 2, \ldots, K\}$. 
The observation $\boldsymbol{X}^{(i)}$ belongs to the Class $k$ if $L^{(i)}=k$ where $i=1, \ldots, n$, and $k=1, 2, \ldots, K$.
Notice that in the semi-supervised learning settings, not all $L^{(i)}$ are available to us.
We denote the observed label by $L_{\ob}^{(i)}$. 
If $i$-th label is missing, then we set $L_{\ob}^{(i)}=0$; otherwise $L_{\ob}^{(i)}=L^{(i)}$.
Here, we assume that the label is missing at random. That is, $\p(L_{\ob}^{(i)} \neq 0|\boldsymbol{X}^{(i)}, L^{(i)})=g(\boldsymbol{X}^{(i)})$.
This assumption makes sense because we do not know the true class $L^{(i)}$ when we decide whether to verify an instance. the only information we have is the observation $\boldsymbol{X}^{(i)}$.

When we talk about a generic observation, we omit the index $i$ from $\boldsymbol{X}^{(i)}$ and just write $\boldsymbol{X}$. We assume that under some unknown increasing transformation $f$, the transformed observations follow one of the $K$ normal distributions according to their classes,
\begin{equation}
    \begin{aligned}
    \boldsymbol{f}(\boldsymbol{X})|\{L=k\} \sim \N_p(\boldsymbol{\mu}_k, \boldsymbol{\Sigma}_k) \quad k \in \{1, 2, \ldots, K \}
\end{aligned}
\end{equation}
where $\boldsymbol{f}$ is a $p$-dimensional vector of functions, $\boldsymbol{f}(X_1, \ldots, X_p)=(f_1(X_1), \ldots, f_p(X_p))$.
Notice that any continuous random variable can be transformed to a normal variable by a strictly increasing transformation, and hence, it is not an assumption if considered individually. The modeling assumption here is that the two distributions for two classes are mapped to normal distributions under the same transformation.
This is called the nonparanormality assumption in the literature \citet{Liu2012}.

The method we use to estimate the transformation $\boldsymbol{f}$ is very similar to that used by \citet{Mulgrave2020}. However, the purpose of estimating this transformation is entirely different. Here, we estimate the transformation $\boldsymbol{f}$ for semi-supervised learning, while \citet{Mulgrave2020} used this approach to learn the graphical structure.
Like \citet{Mulgrave2020}, we shall estimate each component of the transformation $\boldsymbol{f}$. We denote the $d$th dimension of the transformation by $f_d$. 
We put prior distributions on the unknown transformation functions through a random series based on B-splines, i.e., 
\begin{equation*}
    f_d(X_d)=\sum_{j=1}^J \theta_{dj} B_j(X_d), \quad d=1, \ldots, p,
\end{equation*}
where $X_d$ is the $d$th dimension of an observation,  $B_j(\cdot)$ are the B-spline basis functions, $\theta_{dj}$ is a coefficient in the expansion of the function, $j=1, \ldots, J$, and $J$ is the number of B-spline basis functions with equispaced knots used in the expansion. 
The coefficients are ordered to induce monotonicity, and the smoothness is controlled by the order of the B-splines and the number of basis functions used in the expansion. The posterior means of the coefficients give a monotone smooth Bayes estimate of the transformations.
In this paper, we choose cubic splines, which correspond to B-splines of order 4. 

Notice that we can use the B-spline functions to transform the data only if $\boldsymbol{X}$ takes values in $[0,1]$ range. Because of that, we have to transform the data if it is not in this range. For example, we can calculate the mean $\mu_d$ and variance $\sigma^2_d$ for the $d$th dimension of the training data,  and then apply the cumulative distribution function $\Phi(\cdot; \mu_d, \sigma^2_d)$ on the $d$th dimension of the data, $d=1, \ldots, p$. 

\subsection{Prior distributions}

\subsubsection{Prior on the B-spline coefficients:}

The prior we use was introduced by \citet{Mulgrave2020}. Here, we include the prior specifications for completeness.

Let us first put the monotonicity aside, and put normal prior on the coefficients of the B-splines, i.e., $\boldsymbol{\theta}_d=(\theta_{d1}, \ldots, \theta_{dJ}) \sim \N_J(\boldsymbol{\zeta}, \sigma^2 \boldsymbol{I})$, where $\sigma^2$ is some positive constant, $\boldsymbol{\zeta}$ is a $J$ dimensional vector of constants, and $\boldsymbol{I}$ is the $J \times J$ identity matrix. We choose a normal prior for conjugacy. Because the means and the covariances of the normal distributions are unknown, this gives flexibility in the location and the scale of the transformation and causes identifiability issues. To address this and to retain the conjugacy of the normal prior, impose the following constraints on the locations and the scales of the transformation function $f_d$:
\begin{equation}\label{eq:res1}
    \begin{cases}
    0=f_d(1/2)=\sum_{j=1}^J \theta_{dj} B_j(1/2),\\
    1=f_d(3/4)-f_d(1/4)=\sum_{j=1}^J \theta_{dj} [B_j(3/4)-B_j(1/4)].
    \end{cases}
\end{equation}
The constraints can be written in matrix form $\boldsymbol{A}\boldsymbol{\theta}_d=\boldsymbol{c}$, where 
\begin{equation*}
    \boldsymbol{A}=\begin{pmatrix}
    B_1(1/2) & B_2(1/2) & \cdots & B_J(1/2)\\
    B_1(3/4)-B_1(1/4) & B_2(3/4)-B_2(1/4) & \cdots & B_J(3/4)-B_J(1/4)
    \end{pmatrix}
\end{equation*}
and $\boldsymbol{c}=(0,1)^{\mathrm{T}}$.

By the property of the normal distribution, we have
\begin{equation*}
    \boldsymbol{\theta}_d|\{\boldsymbol{A}\boldsymbol{\theta}_d=\boldsymbol{c}\}\sim \N_J(\boldsymbol{\xi}, \boldsymbol{\Gamma}),
\end{equation*}
where $\boldsymbol{\xi}=\boldsymbol{\zeta}+\boldsymbol{A}^{\mathrm{T}}(\boldsymbol{A}\boldsymbol{A}^{\mathrm{T}})^{-1}(\boldsymbol{c}-\boldsymbol{A}\boldsymbol{\zeta})$, $\boldsymbol{\Gamma}=\sigma^2[\boldsymbol{I}-\boldsymbol{A}^{\mathrm{T}}(\boldsymbol{A}\boldsymbol{A}^{\mathrm{T}})^{-1}\boldsymbol{A}]$. Because of the two linear constraints, the covariance matrix $\boldsymbol{\Gamma}$ is actually singular. So, we remove two coefficients by representing them as linear combinations of the others. Here we choose $J_1$ where $B_{J_1}(1/2)$ is the largest (middle one if $J$ is odd, either of middle two if $J$ is even) and $J_2$ where $B_{J_2}(3/4)-B_{J_2}(1/4)$ is the largest (upper $75$th quantile one) and that $J_2 \neq J_1$. Theoretically, any $J_1$ and $J_2$ with non-zero coefficients can work. Here, we choose them to guarantee numerical stability in later calculations. Then by $\boldsymbol{A}\boldsymbol{\theta}_d=\boldsymbol{c}$ we can get
\begin{equation} \label{eq:reduce}
    \begin{pmatrix}
    \theta_{dJ_1}\\
    \theta_{dJ_2}
    \end{pmatrix}
    =\boldsymbol{W}_d\Bar{\boldsymbol{\theta}}_d +\boldsymbol{q}_d,
\end{equation}
where $\Bar{\boldsymbol{\theta}}_d$ is the reduced vector after removing $\theta_{dJ_1}$ and $\theta_{dJ_2}$ from $\boldsymbol{\theta}_d$, and $\boldsymbol{W}_d$ and $\boldsymbol{q}_d$ can be calculated correspondingly.

The reduced vector $\Bar{\boldsymbol{\theta}}_d$ follows the prior distribution:
\begin{equation*}
    \Bar{\boldsymbol{\theta}}_d|\{\boldsymbol{A}\boldsymbol{\theta}_d=\boldsymbol{c}\}\sim \N_{J-2}(\Bar{\boldsymbol{\xi}}, \Bar{\boldsymbol{\Gamma}})
\end{equation*}
where $\Bar{\boldsymbol{\xi}}$ and $\Bar{\boldsymbol{\Gamma}}$ are obtained by removing the $J_1$ and $J_2$ dimension correspondingly.

Finally, consider the monotonicity constraint $\theta_{d1} <\theta_{d2}<\cdots<\theta_{dJ}$. Written in matrix form, that is, $\boldsymbol{F}\boldsymbol{\theta}_d>0$, where
$$\boldsymbol{F}=\begin{pmatrix}
-1 & 1 & 0 & \cdots & 0 & 0\\
0 & -1 & 1 & \cdots & 0 & 0\\
\vdots & \vdots & \vdots & & \vdots & \vdots\\
0 & 0 & 0 & \cdots & -1 & 1
\end{pmatrix}.$$
This can be further reduced to $\Bar{\boldsymbol{F}}\Bar{\boldsymbol{\theta}}_d+\Bar{\boldsymbol{g}}>0$ according to (\ref{eq:reduce}).

The final prior is given by 
\begin{equation} \label{prior:theta}
    \Bar{\boldsymbol{\theta}}_d|\{\boldsymbol{A} \boldsymbol{\theta}_d=c\}\sim \TN_{J-2}(\Bar{\boldsymbol{\xi}}, \Bar{\boldsymbol{\Gamma}}, \mathcal{T})
\end{equation}
where $\mathcal{T}=\{\Bar{\boldsymbol{\theta}}_d: \Bar{\boldsymbol{F}}\Bar{\boldsymbol{\theta}}_d+\Bar{\boldsymbol{g}}>0\}$. Notice that this prior preserves conjugacy.

The parameter $\boldsymbol{\zeta}$ is chosen such that $\zeta_j=\nu+\tau \Phi^{-1} \left( \dfrac{j-0.375}{J-0.75+1} \right), j=1, \ldots, J$, where $\nu$ is a constant and $\tau$ is a positive constant. The motivation behind this prior specification is that this approximates the expected values of the order statistics of a $\N_p(\nu, \tau^2)$ random variable.

\subsubsection{Prior on the means and covariances:}

The means and precision matrices (inverse of the covariance matrix) related to the two Gaussian distributions correspond to unspecified transformed measurements; hence, obtaining prior information on them is difficult. Hence, putting a noninformative prior on them is sensible, i.e., $\pi(\boldsymbol{\mu}_0, \boldsymbol{\Sigma}_0) \propto 1$, $\pi(\boldsymbol{\mu}_1, \boldsymbol{\Sigma}_1) \propto 1$. However, while sampling, a situation can arise when the degrees of freedom of the resulting Wishart posterior distribution of the precision matrix, may be less than the order of the scale matrix. This violates nonsingularity and interrupts the sampler. Thus, to avoid such conditions, we assume the improper uniform prior over the means $\boldsymbol{\mu}_i$ but assume $\boldsymbol{\Sigma}^{-1}_i \sim \W_p(p+2, \boldsymbol{I})$ for $i=0,1$ as the prior on the precision matrices.

\subsubsection{Initializing the B-spline coefficients:}

The monotone nature of the coefficients and that it satisfies $\mathcal{T}$ as given above makes it difficult to assign an initial value to the vector of B-spline coefficients for a given component function. Note that the set $\mathcal{T}$ can be seen as a convex polytope. Thus, if we can provide a solution in the polytope as an initial point, the sampler would continue sampling within $\mathcal{T}$. We assign the initial values for B-spline coefficients $\Bar{\boldsymbol{\theta}}_1, \ldots, \Bar{\boldsymbol{\theta}}_p$ by assuming that the true transformation is $c_0\Phi^{-1}(\cdot)$ for some constant $c_0>0$. These initial values will also be useful when sampling from the posteriors, which are truncated normal distributions. Further, $c_0\Phi^{-1}(\cdot)$ should also satisfy the restrictions (\ref{eq:res1}). The first restriction is trivially satisfied as $\Phi(0) = 1/2$. To meet the second restriction, we set $c_0 = \left(\Phi^{-1}(3/4) - \Phi^{-1}(1/4)\right)^{-1}$. We consider the initial values of $\Bar{\boldsymbol{\theta}}_d$ as the solution $\boldsymbol{\theta}^*$ of the nonlinear least squares problem: 
\[
\boldsymbol{\theta}^* = \underset{\theta \in \mathcal{T}}{\mathrm{minimize}} \int \left(c_0\Phi^{-1}(\boldsymbol{x}) - \boldsymbol{\theta}'\boldsymbol{B}(\boldsymbol{x})\right)^2 d\boldsymbol{x}.
\]
Computation of $\boldsymbol{\theta}^*$ requires theoretically obtaining the inner products $\int B_j(\boldsymbol{x})B_k(\boldsymbol{x})$ of paires of B-spline functions. We can avoid that by computing the minimizer using Monte-Carlo method: obtain a large number of samples $U_1, \cdots, U_N$ independently from the uniform distribution over $(0, 1)$ and compute the least squares solution for $\sum_{i=1}^N \left(c_0\Phi^{-1}(U_i) - \boldsymbol{\theta}'\boldsymbol{B}(U_i)\right)^2$. Such constrained nonlinear least squares problems can be solved efficiently using R packages like \texttt{limSolve} and \texttt{quadprog}. We use \texttt{quadprog} for our numerical studies.

\subsection{Gibbs sampling algorithm}

Denote the transformed observations by $\boldsymbol{Y}$, i.e., $\boldsymbol{Y}_d^{(i)}=f_d(\boldsymbol{X}_d^{(i)})=\boldsymbol{B}_{d}^{(i)} \boldsymbol{\theta}_d$, where $\boldsymbol{B}_{d}^{(i)}=(B_1(X_d^{(i)}), \ldots,$ $B_J(X_d^{(i)}))$, $i=1, \ldots, n$. Because of (\ref{eq:reduce}), we can calculate $\boldsymbol{Y}^{(i)}$ based on $\Bar{\boldsymbol{\theta}}_d$, i.e.,
\begin{equation} \label{eq:Y}
    \boldsymbol{Y}_d^{(i)}=(\Bar{\boldsymbol{B}}_{d}^{(i)}+\boldsymbol{B}_{d}^{(i)*}\boldsymbol{W}_d) \Bar{\boldsymbol{\theta}}_d+\boldsymbol{B}_{d}^{(i)*}\boldsymbol{q}_d,
\end{equation}
where $\Bar{\boldsymbol{B}}_{d}^{(i)}$ is obtained by removing the $J_1$th and the $J_2$th columns of $\boldsymbol{B}_{d}^{(i)}$, and  $\boldsymbol{B}_{d}^{(i)*}$ is given by $(B_{J_1}(X_d^{(i)}), B_{J_2}(X_d^{(i)}))$.

After getting the initial values of the B-spline coefficients using the method defined above, $\Bar{\boldsymbol{\theta}}_d$, we can calculate the initial value for $\boldsymbol{Y}$ according to (\ref{eq:Y}). 
The initial values for $\boldsymbol{\mu}_0$ and $\boldsymbol{\Sigma}_0$ is the mean and the covariance matrix of $\boldsymbol{Y}$ whose $L_{\ob}^{(i)}=0$ respectively; similarly the initial values for $\boldsymbol{\mu}_1$ and $\boldsymbol{\Sigma}_1$ is the mean and the covariance matrix of the $\boldsymbol{Y}$ whose $L_{\ob}^{(i)}=1$.
The original value for the class $\boldsymbol{L}$ is given as follows:
for $i=1, \ldots, n$, 
\begin{equation*}
    L^{(i)}=\begin{cases}
    L_{\ob}^{(i)}, \quad &\mbox{if } L_{\ob}^{(i)}\neq 0,\\
    k, &\mbox{if } L_{\ob}^{(i)}= 0 \mbox{ and } \displaystyle\argmin_{k \in \{1,\ldots, K\}} \lVert \boldsymbol{Y}^{(i)}-\boldsymbol{\mu}_k \rVert =k 
    \end{cases}
\end{equation*}
where $\lVert \cdot \rVert$ stands for the Euclidean distance. 

\begin{enumerate}[leftmargin=*]
    \item First sample the B-spline coefficients for $d=1, \ldots, p$. Let $\boldsymbol{\mu} := \{\boldsymbol{\mu}_1, \ldots, \boldsymbol{\mu}_K\}$ and $\boldsymbol{\Sigma} := \{\boldsymbol{\Sigma}_1, \ldots, \boldsymbol{\Sigma}_K\}$.
    
    The joint posterior for the B-spline coefficients is a truncated normal with density:
    \begin{alignat*}{2}
&\pi^*(\Bar{\boldsymbol{\theta}}_1, \ldots, \Bar{\boldsymbol{\theta}}_p| \boldsymbol{Y}, \boldsymbol{\mu}, \boldsymbol{\Sigma}) \\\nonumber
&\propto \left( \prod_{k=1}^K \prod_{L^{(i)}=0} (\det \boldsymbol{\Sigma}_k)^{-1/2} \exp(-\dfrac{1}{2} (\boldsymbol{Y}^{(i)}-\boldsymbol{\mu}_k)^{\mathrm{T}} \boldsymbol{\Sigma}_k^{-1} (\boldsymbol{Y}^{(i)}-\boldsymbol{\mu}_k))  \right)
\times \left( \prod_{d=1}^p \pi_d(\Bar{\boldsymbol{\theta}}_d) \right)
\end{alignat*}
where $\pi_d(\bar{\theta}_d)$ is the prior distribution of $\bar{\theta}_d$ given by (\ref{prior:theta}).

    Let $\Bar{\boldsymbol{\theta}}_{-d}$ denote the vector $(\Bar{\boldsymbol{\theta}}_1, \ldots, \Bar{\boldsymbol{\theta}}_{d-1}, \Bar{\boldsymbol{\theta}}_{d+1}, \ldots, \Bar{\boldsymbol{\theta}}_{p})$. We have
    \begin{equation}
        \pi^*(\Bar{\boldsymbol{\theta}}_d | \boldsymbol{Y}, \Bar{\boldsymbol{\theta}}_{-d}, \boldsymbol{\mu}_0, \boldsymbol{\Sigma}_0, \boldsymbol{\mu}_1, \boldsymbol{\Sigma}_1) 
        \propto \exp \bigg\{-\dfrac{1}{2} \Bar{\boldsymbol{\theta}}_d^{\mathrm{T}} {\Sigma^*}^{-1} \Bar{\boldsymbol{\theta}}_d+\Bar{\boldsymbol{\theta}}_d^{\mathrm{T}} {\Sigma^*}^{-1}\mu^*\bigg\}  \times \mathbbm{1}\{\Bar{\boldsymbol{F}}\Bar{\boldsymbol{\theta}}_d+\Bar{\boldsymbol{g}}>0\},
    \end{equation}
    So we can sample $\Bar{\boldsymbol{\theta}}_d$ from $\TN(\mu^*, \Sigma^*, \{\Bar{\boldsymbol{F}}\Bar{\boldsymbol{\theta}}_d+\Bar{\boldsymbol{g}}>0\})$,
    where \begin{align*}
        \Sigma^*=&\left\{\sum_{k=1}^{K} \left( (\boldsymbol{\Sigma}_k)^{-1}_{dd} \sum_{L^{(i)}=k}\Big(\Bar{\boldsymbol{B}}_{d}^{(i)}+\boldsymbol{B}_{d}^{(i)*}\boldsymbol{W}_d\Big)^{\mathrm{T}}\Big(\Bar{\boldsymbol{B}}_{d}^{(i)}+\boldsymbol{B}_{d}^{(i)*}\boldsymbol{W}_d\Big)\right)+\Bar{\boldsymbol{\Gamma}}^{-1} \right\}^{-1},\\
        \mu^*=&\Sigma^* \Biggl\{\Bar{\boldsymbol{\xi}}\Bar{\boldsymbol{\Gamma}}^{-1}- \sum_{k=1}^K \sum_{L^{(i)}=k}\Big(\Bar{\boldsymbol{B}}_{d}^{(i)}+\boldsymbol{B}_{d}^{(i)*}\boldsymbol{W}_d\Big) \Big[(\boldsymbol{\Sigma}_k)^{-1}_{dd}\Big(\boldsymbol{B}_{d}^{(i)*}\boldsymbol{q}-\boldsymbol{\mu}_{kd}\Big)+ \\ 
        & \qquad\qquad\qquad\qquad\qquad (\boldsymbol{\Sigma}_k)^{-1}_{d,-d}\Big(\boldsymbol{Y}^{(i)}_{-d}-\boldsymbol{\mu}_{k, -d}\Big)\Big] \Biggl\}
    \end{align*}
    The truncated normal distributions are sampled using the method proposed by \citet{Li2015}. After obtaining the posterior samples of $\Bar{\boldsymbol{\theta}}_1, \ldots, \Bar{\boldsymbol{\theta}}_p$, we can update $\boldsymbol{Y}$ according to (\ref{eq:Y}).
    
    \item Update $\boldsymbol{\mu}_0, \boldsymbol{\Sigma}_0, \boldsymbol{\mu}_1, \boldsymbol{\Sigma}_1$ according to their posterior probability. That is,
    \begin{equation*}
        \begin{aligned}
        \boldsymbol{\Sigma}_k^{-1} &\sim \W\left(p+2+n_k, \left(\sum_{L^{(i)}=k}(\boldsymbol{Y}^{(i)}-\Bar{\boldsymbol{Y}}_{k \cdot})(\boldsymbol{Y}^{(i)}-\Bar{\boldsymbol{Y}}_{k \cdot})^{\mathrm{T}} + \boldsymbol{I}\right)^{-1}\right),\\
        \boldsymbol{\mu}_k &\sim \N_p(\Bar{\boldsymbol{Y}}_{k \cdot}, \boldsymbol{\Sigma}_k/n_k), \quad\quad k \in \{1,2,\ldots, K\},
    \end{aligned}
    \end{equation*}
    where $n_k=\sum_{i=1}^n \mathbbm{1}\{L^{(i)}=k\}$, $\Bar{\boldsymbol{Y}}_{k \cdot}=\sum_{i=1}^n \mathbbm{1} \{L^{(i)}=k\} \boldsymbol{Y}^{(i)}/n_k$. 
    
    \item Update the missing labels according to the current distributions; 
    for $i=1, \ldots, n$, if label is missing, update 
    $$L^{(i)} \sim \mathrm{Multinomial} \left(1; \dfrac{\lambda_1 \phi(Y^{(i)}; \boldsymbol{\mu}_1, \boldsymbol{\Sigma}_1)}{\sum_{k=1}^K \lambda_k \phi(Y^{(i)}; \boldsymbol{\mu}_k, \boldsymbol{\Sigma}_k)}, \cdots, \dfrac{\lambda_K \phi(Y^{(i)}; \boldsymbol{\mu}_K, \boldsymbol{\Sigma}_K)}{\sum_{k=1}^K \lambda_k \phi(Y^{(i)}; \boldsymbol{\mu}_k, \boldsymbol{\Sigma}_k)} \right),$$
    since $\dfrac{\p(L^{(i)}=r|Y^{(i)})}{\p(L^{(i)}=s|Y^{(i)})}=\dfrac{\lambda_r \phi(Y^{(i)}; \boldsymbol{\mu}_r, \boldsymbol{\Sigma}_r)}{\lambda_s \phi(Y^{(i)}; \boldsymbol{\mu}_s, \boldsymbol{\Sigma}_s)}$ based on the missing at random assumption, $r,s \in \{1, \ldots, K\}$,
    where $\lambda_r$ and $\lambda_s$ stand for the proportion of class $r$ and class $s$. 
    
    There are two ways to figure out $\boldsymbol{\lambda} = (\lambda_1, \ldots, \lambda_K)'$. One is to specify them in advance if we know the proportion of each category in the population. The other method is to treat them as unknown parameters. We can specify the prior distribution $\boldsymbol{\lambda} \sim \mathrm{Dirichlet}(l_1, \ldots,  l_K)$ and update them with each MCMC iteration $\boldsymbol{\lambda} \sim \mathrm{Dirichlet}(l_1+n_1, \ldots, l_K+n_K)$. The numerical experiments in the next two sections are carried out using the latter method.
    
\end{enumerate}
We can then obtain the posterior mean of $\Bar{\boldsymbol{\theta}}_1, \ldots, \Bar{\boldsymbol{\theta}}_p, \Bar{\boldsymbol{\mu}}, \Bar{\boldsymbol{\Sigma}}$ (and $\Bar{\boldsymbol{\lambda}}$ if treated as an unknown parameter). To predict labels for new data points, we proceed as:
\begin{itemize}
    \item For the new data $X^{(\new)}$ coming in, apply the transformation given by the B-spline series with coefficients $\Bar{\boldsymbol{\theta}}_1, \ldots, \Bar{\boldsymbol{\theta}}_p$.
    \item Calculate $Y^{(\new)}$ according to (\ref{eq:Y})
    \item If $\displaystyle\argmax_{k \in \{1, \ldots, K\}}$ $\Bar{\lambda}_k \phi(Y^{(\new)}, \Bar{\boldsymbol{\mu}}_k, \Bar{\boldsymbol{\Sigma}}_k) = k$, then assign it to Class k. 
\end{itemize}

We will call our method the Nonparanormal method in the simulations.

\section{Model selection} \label{sec:select}

We must choose the number of basis functions $J$ used to estimate the transformation function. If the labeled data is sufficiently extensive, we can consider using cross-validation to choose $J$. However, in semi-supervised learning settings, the number of labeled units in the data is usually very limited. Our proposed method is inspired by the low-density assumption widely used in the semi-supervised learning literature. We choose the classifier that best satisfies the low-density assumption, i.e., the one with the least number of points on the boundary. More specifically, we define the points that are close to the boundary by 

\[\displaystyle\bigcup_{k=1}^K \displaystyle\bigcup_{l>k} \left\{X^{(i)}: m^{-1}<\dfrac{\lambda_l \phi(Y^{(i)}; \Bar{\boldsymbol{\mu}_l}, \Bar{\boldsymbol{\Sigma}_l})}{\lambda_k \phi(Y^{(i)}; , \Bar{\boldsymbol{\mu}_k}, \Bar{\boldsymbol{\Sigma}_k})}<m, i=1, \ldots, n \right\}, \]
where  $\Bar{\boldsymbol{\mu}_l}, \Bar{\boldsymbol{\Sigma}_l}, \Bar{\boldsymbol{\mu}_k}, \Bar{\boldsymbol{\Sigma}_k}$ are the posterior means of $\boldsymbol{\mu}_l, \boldsymbol{\Sigma}_l, \boldsymbol{\mu}_k, \boldsymbol{\Sigma}_k$ respectively and $\{k , l\} \in \{1,\ldots, K\}$, $ k < l$. For a binary classification problem, this becomes:
\begin{equation} \label{boundaryset}
\left\{X^{(i)}: m^{-1}<\dfrac{\lambda_0 \phi(Y^{(i)}; \Bar{\boldsymbol{\mu}_0}, \Bar{\boldsymbol{\Sigma}_0})}{\lambda_1 \phi(Y^{(i)}; , \Bar{\boldsymbol{\mu}_1}, \Bar{\boldsymbol{\Sigma}_1})}<m, i=1, \ldots, n \right\}.
\end{equation}
The quantity $m(>1)$ takes a fixed value we need to specify.
The set is more precise to the boundary when $m$ is closer to 1. However, this may lead to a small number of points in the boundary set and, therefore, cannot distinguish different $J$. According to our numerical experiment, moderate value $m=3$ is a good choice.
Here $\boldsymbol{Y}$ is calculated based on the posterior mean $\Bar{\boldsymbol{\theta}}_1, \ldots, \Bar{\boldsymbol{\theta}}_p$. We choose the value of $J$ that has the fewest points in the set defined by (\ref{boundaryset}).

In the simulation studies, to save computation time, we run the procedure with $J=8, \ldots, 15$, each for $500$ iterations to get a decision rule. Then, we choose the best $J$ according to the low-density assumption introduced above and then run $10000$ iterations to reject the first $2000$ samples as burn-in to get the final classifier.

\section{Binary Classification: Simulation Study} 
\label{sec:binsim}

Since our proposed method relies on the nonparanormality assumption, in the simulation studies, we consider two cases: the assumption is satisfied, and the assumption is violated.

\subsection{Nonparanormality assumption satisfied}

For a given dimension $p$, instead of specifying some values for means and covariances of the underlying Gaussian distributions $\N_p(\boldsymbol{\mu}_0, \boldsymbol{\Sigma}_0)$ and $\N_p(\boldsymbol{\mu}_1, \boldsymbol{\Sigma}_1)$, we decide to randomly generate some values as follows:
\begin{itemize}
    \item Generate $\mu_{0d}$ and $\mu_{1d}$ independently and identically from a uniform distribution ranging in $[0,4]$ for $d=1, \ldots, p$.
    \item Generate $\boldsymbol{\Sigma}_0$ by $\boldsymbol{A}^{\mathrm{T}}\boldsymbol{A}$, where every element in $\boldsymbol{A}$ is generated from the uniform distribution over $[-1,1]$, $\boldsymbol{\Sigma}_1$ is generated the same way independently.
\end{itemize}

Here, we considered two true transformations: 
\begin{enumerate}
    \item The \textit{Logistic} transformation: $\left\{1+\exp\left(-\frac{X_d-(\mu_{0d}+\mu_{1d})/2}{(\sqrt{\Sigma_{0 d,d}}+\sqrt{\Sigma_{1 d,d}})/2}\right)\right\}^{-1}$ for $d=1, \ldots, p$
    \item The \textit{Gumbel} transformation: $\exp\left\{-\exp\left(-\frac{X_d-(\mu_{0d}+\mu_{1d})/2}{(\sqrt{\Sigma_{0 d,d}}+\sqrt{\Sigma_{1 d,d}})/2}\right)\right\}$ for $d=1, \ldots, p$ 
\end{enumerate}

For each of the settings above, we take $p$ ranging between $5$ and $100$ and generate $n$ observations between $100$ and $500$. We then split $n$ by a $20:80$, $30:70$, or $40:60$ split and assign the respective parts to $n_1$, the number of observations in class 0, and $n_2$, the number of observations in class 1. We select $n_L$ in the range $3$ to $20$, where $n_L$ is the number of labeled data points in each class. We generate $4000$ data points from class 0 and $6000$ from class 1 to test the model. 

\begin{table}[!b]
\centering
\caption{Classification error percentage ((FP+FN)/2 $\times 100\%$) for the test data when the data is generated with a logistic transformation on both classes.}
\resizebox{0.95\textwidth}{!}{
\begin{tabular}{@{}llllllllllll@{}}
\toprule 
  $\mathbf{p}$ &
  $\mathbf{n_1}$ &
  $\mathbf{n_2}$ &
  $\mathbf{n_L}$ &
  \textbf{NPN} &
  \textbf{EMLS} &
  \textbf{ICLS} &
  \textbf{LaSVM} &
  \textbf{LR} &
  \textbf{WSVM} &
  \textbf{SVML} &
  \textbf{EMNM} \\ \midrule
5   & 20  & 80  & 3  & 7.37042 & 13.02 & 15.12 & 26.92 & 25.86 & 27.21 & 27.66 & 45.98 \\
5   & 30  & 70  & 3  & 6.29903 & 14.03 & 14.23 & 29.12 & 28.6  & 23.44 & 29    & 44.33 \\
5   & 40  & 60  & 3  & 5.76431 & 16    & 16.5  & 30.84 & 29.33 & 24.36 & 31.07 & 43.58 \\
10  & 40  & 60  & 3  & 0.77667 & 14.49 & 19.93 & 15.05 & 14.06 & 9     & 13.28 & 8.27  \\
10  & 60  & 240 & 5  & 0.31444 & 8.17  & 12.64 & 11.36 & 11.07 & 18.31 & 10.59 & 11.62 \\
15  & 90  & 210 & 5  & 0.06569 & 9.23  & 16.92 & 9.21  & 12.22 & 11.19 & 9.77  & 4.85  \\
30  & 120 & 180 & 5  & 0.01333 & 31.96 & 31.96 & 22.13 & 23.7  & 14.53 & 21.39 & 16.49 \\
50  & 120 & 180 & 5  & 0       & 26.37 & 35.08 & 24.37 & 28.22 & 13.58 & 23.35 & 17.47 \\
25  & 200 & 300 & 7  & 0.01014 & 16.08 & 23.17 & 13.89 & 14.47 & 6.9   & 12.7  & 7.47  \\
25  & 150 & 350 & 7  & 0.01542 & 14.88 & 21.41 & 14.67 & 14.69 & 12.97 & 13.16 & 8.21  \\
25  & 100 & 400 & 7  & 0.01694 & 17.3  & 23.77 & 13.72 & 14.6  & 18.23 & 12.78 & 15.08 \\
50  & 200 & 300 & 7  & 0       & 31.63 & 32.74 & 21.45 & 23.42 & 9.66  & 20.35 & 13.23 \\
50  & 150 & 350 & 7  & 0       & 30.57 & 32.81 & 22.59 & 24.35 & 16.75 & 21.65 & 23.22 \\
50  & 100 & 400 & 7  & 0.00125 & 32.24 & 33.74 & 22.28 & 24.45 & 22.05 & 21.64 & 36.25 \\
50  & 200 & 300 & 10 & 0       & 24.68 & 27.65 & 17.64 & 18.65 & 8.56  & 16.54 & 12.36 \\
50  & 150 & 350 & 10 & 0       & 25.7  & 28.48 & 18.35 & 19.16 & 14.72 & 17.35 & 19.95 \\
50  & 100 & 400 & 10 & 0.00125 & 26.88 & 29.77 & 18.49 & 18.9  & 19.78 & 17.47 & 32.84 \\
50  & 400 & 600 & 12 & 0       & 19.75 & 26.69 & 15.99 & 16.57 & 7.14  & 14.79 & 10.83 \\
50  & 300 & 700 & 12 & 0       & 20.42 & 27.01 & 16.52 & 17.15 & 13.91 & 15.62 & 17.6  \\
50  & 200 & 800 & 12 & 0.00042 & 21.67 & 27.12 & 16.34 & 17.35 & 19.52 & 15.71 & 32.9  \\
75  & 400 & 600 & 15 & 0       & 26.23 & 30.13 & 19.76 & 22.52 & 10.72 & 19.51 & 13.91 \\
75  & 300 & 700 & 15 & 0       & 27.24 & 29.53 & 18.64 & 20.38 & 15.34 & 18.03 & 17.95 \\
75  & 200 & 800 & 15 & 0       & 26.23 & 29.38 & 18.18 & 20.37 & 20.12 & 17.77 & 27.06 \\
100 & 400 & 600 & 20 & 0       & 26.36 & 29.53 & 19.64 & 22.07 & 10.91 & 19.18 & 22.92 \\
100 & 300 & 700 & 20 & 0       & 29.97 & 31.02 & 21.26 & 24.45 & 16.43 & 20.74 & 27.74 \\
100 & 200 & 800 & 20 & 0       & 28.58 & 30.38 & 20.31 & 23.03 & 20.94 & 19.99 & 35.12 \\ \bottomrule
\end{tabular}
}
\label{tab:ll_simul}
\end{table}

We compare our Nonparanormal(NPN) method to other widely used semi-supervised learning methods in R package \texttt{RSSL} of \citet{Krijthe2017}. We have tried all the methods included in this package. However, EM Linear Discriminant Classifier (Expectation Maximization applied to the linear discriminant classifier assuming Gaussian classes with a shared covariance matrix) \cite{Dempster1977}, IC Linear Discriminant Classifier  \cite{Krijthe2014}, Kernel Least Squares Classifier, Laplacian Kernel Least Squares Classifier \cite{Belkin2006}, 
MC Linear Discriminant Classifier \cite{Loog2011} and Quadratic Discriminant Classifier do not work for our generated datasets because the R code throws an error upon compilation. Entropy Regularized Logistic Regression  \cite{Grandvalet2005}, Linear SVM, Logistic Loss Classifier, Logistic Regression, MC Nearest Mean Classifier \cite{Loog2010}, and EM Least Squares classifier have average and comparable performances. Our study includes the classical Logistic Regression and EM-based Least Squares methods. Other than these, we include 5 other methods that perform relatively better on our simulated data. These methods are IC Least Squares Classifier (ICLS) \cite{Krijthe2015}, Laplacian SVM (LSVM) \cite{Belkin2006}, Well SVM (WSVM) \cite{Li2013}, svmlin (SVML) \cite{Sindhwani2006}, and EM Nearest Mean Classifier (EM) \cite{Dempster1977}. We simulate 30 datasets for each setting, and then for each of them, we record their false positive (FP) and false negative (FN) error rates. The results tabulated in Tables \ref{tab:ll_simul} and \ref{tab:gg_simul} are presented in percentages of the combined quantity as (FP+FN)/2 $\times 100\%$.


\begin{table}[t]
\centering
\caption{Classification error percentage ((FP+FN)/2 $\times 100\%$) for the test data when the data is generated with a Gumbel transformation on both classes.}
\resizebox{0.95\textwidth}{!}{
\begin{tabular}{@{}llllllllllll@{}}
\toprule
  $\mathbf{p}$ &
  $\mathbf{n_1}$ &
  $\mathbf{n_2}$ &
  $\mathbf{n_L}$ &
  \textbf{NPN} &
  \textbf{EMLS} &
  \textbf{ICLS} &
  \textbf{LaSVM} &
  \textbf{LR} &
  \textbf{WSVM} &
  \textbf{SVML} &
  \textbf{EMNM} \\ \midrule
5   & 20  & 80  & 3  & 11.24972 & 17.15 & 19.33 & 26.13 & 25.5  & 27.85 & 26.13 & 46.19 \\ 
5   & 30  & 70  & 3  & 9.25681  & 18.27 & 18.34 & 29.88 & 27.75 & 25.11 & 28.05 & 45.13 \\
5   & 40  & 60  & 3  & 9.20292  & 19.52 & 20.59 & 31.6  & 28.33 & 25.85 & 29.74 & 44.14 \\
10  & 40  & 60  & 3  & 0.75847  & 17.29 & 21.73 & 15.58 & 14.28 & 9.68  & 13.45 & 9.07  \\
10  & 60  & 240 & 5  & 0.55139  & 11.78 & 14.74 & 11.79 & 11.74 & 18.43 & 10.89 & 12.95 \\
15  & 90  & 210 & 5  & 0.08389  & 11.24 & 17.5  & 10.04 & 12.1  & 11.54 & 10.06 & 5.23  \\
30  & 120 & 180 & 5  & 0.00861  & 32.16 & 32.47 & 22.77 & 24.06 & 15.56 & 22.07 & 17.28 \\
50  & 120 & 180 & 5  & 0.00194  & 27.47 & 35.13 & 25.05 & 25.52 & 14.68 & 24.24 & 17.73 \\
25  & 200 & 300 & 7  & 0.01514  & 18.01 & 24.49 & 14.58 & 15.35 & 7.55  & 13.67 & 7.92  \\
25  & 150 & 350 & 7  & 0.02222  & 17.26 & 22.87 & 15.08 & 15.06 & 13.63 & 13.98 & 8.59  \\
25  & 100 & 400 & 7  & 0.02972  & 19.1  & 24.65 & 14.33 & 14.89 & 18.51 & 13.61 & 15.44 \\
50  & 200 & 300 & 7  & 0        & 32.65 & 33.71 & 22.53 & 23.44 & 10.87 & 21.66 & 14.44 \\
50  & 150 & 350 & 7  & 0.00042  & 31.97 & 33.32 & 23.44 & 24.25 & 17.68 & 22.7  & 24.36 \\
50  & 100 & 400 & 7  & 0.00792  & 33.53 & 34.4  & 23.03 & 24.29 & 22.77 & 22.53 & 36.08 \\
50  & 200 & 300 & 10 & 0.00028  & 26.65 & 28.98 & 18.71 & 21.43 & 9.62  & 17.88 & 14    \\
50  & 150 & 350 & 10 & 0.0025   & 27.04 & 29.06 & 19.32 & 21.36 & 15.42 & 18.46 & 23.15 \\
50  & 100 & 400 & 10 & 0.00583  & 28.01 & 30.65 & 19.43 & 21.48 & 20.33 & 18.71 & 33.04 \\
50  & 400 & 600 & 12 & 0        & 22.15 & 27.93 & 16.93 & 19.31 & 7.83  & 16.3  & 12.42 \\
50  & 300 & 700 & 12 & 0.00042  & 23.08 & 27.9  & 17.58 & 19.63 & 14.43 & 16.92 & 18.99 \\
50  & 200 & 800 & 12 & 0.00167  & 23.82 & 28.12 & 17.24 & 19.53 & 19.98 & 17.02 & 32.22 \\
75  & 400 & 600 & 15 & 0        & 28.26 & 31.03 & 20.57 & 25.49 & 11.43 & 20.69 & 14.73 \\
75  & 300 & 700 & 15 & 0        & 29.12 & 30.46 & 19.5  & 23.41 & 15.98 & 19.23 & 18.24 \\
75  & 200 & 800 & 15 & 0        & 27.55 & 29.85 & 19.01 & 23.56 & 20.51 & 19.07 & 27.13 \\
100 & 400 & 600 & 20 & 0        & 27.49 & 30.12 & 20.16 & 24.56 & 11.35 & 19.98 & 23.06 \\
100 & 300 & 700 & 20 & 0        & 30.88 & 31.5  & 21.87 & 26.18 & 17    & 21.71 & 27.7  \\
100 & 200 & 800 & 20 & 0        & 29.69 & 31.03 & 20.93 & 25.5  & 21.09 & 20.9  & 35.18 \\ \bottomrule
\end{tabular}
}
\label{tab:gg_simul}
\end{table}

The simulation results clearly show that NPN outperforms other methods when our assumption is satisfied. 
It is worth pointing out that because the parameters generated for different dimensions are different, the classification difficulty is also different. The difficulty for classification decreases as the dimensions increase in our setting. The accuracy increases as the number of labels increases, which is expected. The accuracy also increases as the number of samples and the dimension of the data increases. For NPN, more data in terms of dimensions and sample size and more labeled data per class proves efficient as the model can more easily distinguish between the two class distributions. In some such cases, exact recovery of the cluster memberships is possible, and this seems reasonable as the simulated data do not contain any contaminated observations.
Comparing different methods, the EM algorithm is highly dependent on the assumption that the underlying distributions are Gaussian. This assumption is not applicable in these cases; thus, EM-based methods do not perform well. Among the SVM-based methods, all three have comparable and competitive performances. These methods did not do well in our cases because of the low-density assumption of the decision boundary that did not satisfy our cases. Overall, the noticeably larger classification error rates of the methods compared to NPN may result from the fact that those methods could not bypass the effect of the distributional transformations in the data.

\subsection{Nonparanormality assumption fails}

\begin{table}[!b]
\centering
\caption{Classification error percentage ((FP+FN)/2 $\times 100\%$) for the test data when the data violate the Nonparanormal assumption.}
\resizebox{0.95\textwidth}{!}{
\begin{tabular}{@{}llllllllllll@{}}
\toprule
  $\mathbf{p}$ &
  $\mathbf{n_1}$ &
  $\mathbf{n_2}$ &
  $\mathbf{n_L}$ &
  \textbf{NPN} &
  \textbf{EMLS} &
  \textbf{ICLS} &
  \textbf{LaSVM} &
  \textbf{LR} &
  \textbf{WSVM} &
  \textbf{SVML} &
  \textbf{EMNM} \\ \midrule
5   & 20  & 80  & 3  & 1.45833 & 6.81  & 9.3   & 11.01 & 14.26 & 18.39 & 16.42 & 31.31 \\ 
5   & 30  & 70  & 3  & 0.25444 & 6.96  & 7.82  & 13.41 & 15.25 & 13.3  & 17.95 & 20.6  \\
5   & 40  & 60  & 3  & 0.23431 & 7.55  & 10.03 & 14.03 & 15.56 & 7.56  & 18.18 & 14.1  \\
10  & 40  & 60  & 3  & 0.05181 & 8.96  & 16.38 & 10.15 & 11.62 & 6.05  & 11.59 & 4.81  \\
10  & 60  & 240 & 5  & 0.00694 & 5.11  & 8.65  & 6.9   & 9.52  & 16.48 & 9.16  & 8.42  \\
15  & 90  & 210 & 5  & 0.16819 & 15.59 & 20    & 12.87 & 13.47 & 12.24 & 11.07 & 9.54  \\
30  & 120 & 180 & 5  & 0.00097 & 22.58 & 26.87 & 10.24 & 21.38 & 5.13  & 18.52 & 3.15  \\
50  & 120 & 180 & 5  & 0       & 17.21 & 31.99 & 10.69 & 28.84 & 3.73  & 22.33 & 0.97  \\
25  & 200 & 300 & 7  & 0.00306 & 12.15 & 20.07 & 9.85  & 19.09 & 5.25  & 12.41 & 4.77  \\
25  & 150 & 350 & 7  & 0.00153 & 11.46 & 19.18 & 9.45  & 17.93 & 10.56 & 11.84 & 5.02  \\
25  & 100 & 400 & 7  & 0.00375 & 14.65 & 21.85 & 9.17  & 17.87 & 15.79 & 12.42 & 6.88  \\
50  & 200 & 300 & 7  & 0       & 21.78 & 28.97 & 7.28  & 24.62 & 2.92  & 18.49 & 0.82  \\
50  & 150 & 350 & 7  & 0       & 21.72 & 28.11 & 8.36  & 24.54 & 9.59  & 20.15 & 0.85  \\
50  & 100 & 400 & 7  & 0.00083 & 21.19 & 27.71 & 7.27  & 24.01 & 16.26 & 19.64 & 1.54  \\
50  & 200 & 300 & 10 & 0       & 13.34 & 23    & 4.11  & 19.76 & 2.66  & 14.14 & 0.84  \\
50  & 150 & 350 & 10 & 0       & 13.72 & 22.12 & 5.05  & 19.8  & 8.69  & 15.8  & 0.86  \\
50  & 100 & 400 & 10 & 0.00042 & 14.39 & 22.01 & 4.54  & 20    & 15.41 & 15.4  & 1.49  \\
50  & 400 & 600 & 12 & 0       & 6.54  & 18.77 & 3.68  & 18.02 & 2.99  & 13.46 & 0.76  \\
50  & 300 & 700 & 12 & 0       & 7.85  & 18.45 & 4.01  & 17.9  & 10.06 & 13.92 & 0.74  \\
50  & 200 & 800 & 12 & 0       & 8.28  & 19.56 & 3.55  & 17.57 & 16.36 & 13.5  & 0.88  \\
75  & 400 & 600 & 15 & 0       & 7.67  & 22.12 & 2.71  & 21.37 & 2.69  & 16.56 & 0.69  \\
75  & 300 & 700 & 15 & 0       & 7.69  & 21.17 & 2.29  & 20.54 & 8.12  & 15.38 & 0.62  \\
75  & 200 & 800 & 15 & 0       & 8.81  & 20.35 & 2.26  & 19.89 & 14.25 & 15.6  & 0.58  \\
100 & 400 & 600 & 20 & 0       & 9.88  & 22.91 & 1.61  & 19.36 & 1.75  & 15.84 & 0.4   \\
100 & 300 & 700 & 20 & 0       & 12.08 & 22.94 & 1.59  & 21.69 & 6.58  & 17.03 & 0.33  \\
100 & 200 & 800 & 20 & 0       & 10.92 & 21.18 & 1.57  & 20.17 & 12.71 & 16.78 & 0.27 \\ \bottomrule 
\end{tabular}
}
\label{tab:lg_simul}
\end{table}

To simulate the case when the nonparanormality assumption is violated, we still generate Gaussian distributions  $\N_p(\boldsymbol{\mu}_0, \boldsymbol{\Sigma}_0)$ and $\N_p(\boldsymbol{\mu}_1, \boldsymbol{\Sigma}_1)$ as the underlying distributions, but we will apply different transformation on different classes. We will use the same values generated above for means and covariances for simplicity. 
The transformation we applied are:
\begin{enumerate}
    \item The \textit{Logistic} transformation: $\left\{1+\exp\left(-\frac{X_d-(\mu_{0d}+\mu_{1d})/2}{(\sqrt{\Sigma_{0 d,d}}+\sqrt{\Sigma_{1 d,d}})/2}\right)\right\}^{-1}$ for $d=1, \ldots, p$ on observations of Class 0.
    \item The \textit{Gumbel} transformation: $\exp\left\{-\exp\left(-\frac{X_d-(\mu_{0d}+\mu_{1d})/2}{(\sqrt{\Sigma_{0 d,d}}+\sqrt{\Sigma_{1 d,d}})/2}\right)\right\}$ for $d=1, \ldots, p$ on observations of Class 1. 
\end{enumerate}
Further, the choices of sample sizes, $p$, $n_L$, and the test set size are the same as above. Again, 30 datasets are generated for each setting, and we compare our proposed method with the abovementioned methods. The results are tabulated in Table \ref{tab:lg_simul}. The results show that the proposed NPN method has substantially lower error rates than others, even when the nonparanormality assumption fails. Thus, our method is robust to this assumption.

\section{Binary Classification: Real Data Analysis} 
\label{sec:binreal}

For the real data part, we follow the literature by choosing datasets with labels and then generating missing labels so that we can compare the performances of different methods.
Here, we consider two datasets from UCI Machine Learning Repository\footnote{https://archive.ics.uci.edu/ml/index.php}. For each dataset, we randomly select 70\% of the data for training purposes and 30\% of the data as a testing set. For the training set, we again randomly select 15\% of the label and regard the rest as unlabeled data for our semi-supervised learning method. For each dataset, we repeat the process 10 times and take the mean of the false positive rate, the false negative rate, and the Matthews correlation coefficient (MCC) for all semi-supervised methods considered. We also tabulate the standard error under its corresponding estimate. Given TP as the number of true positives, FP as false positives, TN as true negatives, and FN as false negatives, MCC is given by:
\[
\text{MCC} = \frac{\mathrm{TP} \times \mathrm{TN} - \mathrm{FP} \times \mathrm{FN}}{\sqrt{\mathrm{(TP + FP)(TP + FN)(TN + FP)(TN + FN)}}}
\]
The higher the value of MCC, better is the performance of the classifier. We calculate the MCC because it considers true and false positives and negatives and is generally regarded as a balanced measure that can be used even if the classes are of very different sizes.
The two data sets we considered are:

\begin{itemize}
    \item Breast Cancer Wisconsin (Diagnostic) Data Set \footnote{https://archive.ics.uci.edu/ml/datasets/Breast+Cancer+Wisconsin+(Diagnostic)}
    
    The purpose of this data is to use the features of the cell nucleus to predict whether the disease is malignant or benign. The features are computed from a digitized image of a breast mass's fine needle aspirate (FNA). Those features include radius, texture, perimeter, area, smoothness, compactness, concavity, concave points, symmetry, and fractal dimension. Each feature's mean, standard error, and ``worst'' or largest (mean of the three largest values) are calculated for each image. All features are continuous variables. The data have 357 benign cases and 212 malignant cases.
    
    \item Ionosphere Data Set
    \footnote{https://archive.ics.uci.edu/ml/datasets/ionosphere}
    
    This data aims to classify the radar returns of free electrons in the ionosphere as ``good'' or ``bad'': ``Good'' radar returns show evidence of some type of structure in the ionosphere, while ``bad'' returns are those whose signals pass through the ionosphere. The system has 17 pulse numbers, and instances are described by 2 attributes per pulse number. This data set has 34 features, and all of them are continuous. The data contains 126 bad cases and 225 good cases.
\end{itemize}

\begin{table}[t]
\centering
\caption{Classification results for \texttt{Breast Cancer Wisconsin (Diagnostic)} Dataset}
\resizebox{1.0\textwidth}{!}{
\begin{tabular}{@{}llrrrrrrr@{}}
\toprule
  \textbf{} &
  \textbf{NPN} &
  \multicolumn{1}{l}{\textbf{EMLS}} &
  \multicolumn{1}{l}{\textbf{ICLS}} &
  \multicolumn{1}{l}{\textbf{LaSVM}} &
  \multicolumn{1}{l}{\textbf{LR}} &
  \multicolumn{1}{l}{\textbf{WSVM}} &
  \multicolumn{1}{l}{\textbf{SVML}} &
  \multicolumn{1}{l}{\textbf{EMNM}} \\ \midrule 
\textbf{False Positive Rate} & \textbf{0.032} & 0.318 & 0.269 & 0.185 & 0.139 & 0.194 & 0.109 & 0.367 \\
\textbf{}                    & ($\pm$0.018) & ($\pm$0.084) & ($\pm$0.086) & ($\pm$0.075) & ($\pm$0.046) & ($\pm$0.059) & ($\pm$0.065) & ($\pm$0.058) \\ \hline 
\textbf{False Negative Rate} & 0.196 & 0.021 & 0.023 & \textbf{0.007} & 0.071 & 0.037 & 0.085 & 0.002 \\
\textbf{}                    & ($\pm$0.084) & ($\pm$0.017) & ($\pm$0.022) & ($\pm$0.011) & ($\pm$0.061) & ($\pm$0.036) & ($\pm$0.065) & ($\pm$0.004) \\ \hline
\textbf{MCC}                 & 0.799 & 0.723 & 0.757 & \textbf{0.846} & 0.796 & 0.797 & 0.806 & 0.717 \\
\textbf{}                    & ($\pm$0.059) & ($\pm$0.067) & ($\pm$0.079) & ($\pm$0.049) & ($\pm$0.066) & ($\pm$0.047) & ($\pm$0.063) & ($\pm$0.045) \\ \bottomrule
\end{tabular}
}
\label{tab:wdbc}
\end{table}

\begin{table}[t]
\centering
\caption{Classification results for \texttt{Ionosphere} Dataset}
\resizebox{1.0\textwidth}{!}{
\begin{tabular}{@{}llrrrrrrr@{}} \toprule
\multicolumn{1}{l}{} &
  \textbf{NPN} &
  \multicolumn{1}{l}{\textbf{EMLS}} &
  \multicolumn{1}{l}{\textbf{ICLS}} &
  \multicolumn{1}{l}{\textbf{LaSVM}} &
  \multicolumn{1}{l}{\textbf{LR}} &
  \multicolumn{1}{l}{\textbf{WSVM}} &
  \multicolumn{1}{l}{\textbf{SVML}} &
  \multicolumn{1}{l}{\textbf{EMNM}} \\ \midrule
\textbf{False Positive Rate} & 0.116 & \textbf{0.032} & 0.053 & 0.058 & 0.077 & 0.117 & 0.038 & 0.305 \\
\textbf{}                    & ($\pm$0.044) & ($\pm$0.047) & ($\pm$0.063) & ($\pm$0.136) & ($\pm$0.065) & ($\pm$0.130) & ($\pm$0.041) & ($\pm$0.059) \\ \hline
\textbf{False Negative Rate} & \textbf{0.119} & 0.526 & 0.490 & 0.584 & 0.447 & 0.330 & 0.500 & 0.267 \\
\textbf{}                    & ($\pm$0.033) & ($\pm$0.076) & ($\pm$0.081) & ($\pm$0.208) & ($\pm$0.114) & ($\pm$0.082) & ($\pm$0.105) & ($\pm$0.053) \\ \hline
\textbf{MCC}                 & \textbf{0.751} & 0.545 & 0.541 & 0.473 & 0.535 & 0.591 & 0.555 & 0.415 \\
\textbf{}                    & ($\pm$0.049) & ($\pm$0.065)) & ($\pm$0.101) & ($\pm$0.116) & ($\pm$0.116) & ($\pm$0.133) & ($\pm$0.075) & ($\pm$0.078) \\ \bottomrule
\end{tabular}
}
\label{tab:ionos}
\end{table}

The results are presented in Table \ref{tab:wdbc} and Table \ref{tab:ionos}. For the \textit{Breast Cancer} data, LaSVM has the best overall error rate and MCC. Our proposed method, NPN, has slightly higher error rates and lower MCC. However, unlike LaSVM, whose false negative rate is much higher than the false positive rate, the false positive rate and false negative rate of NPN are almost the same. WSVM has error rates and MCC that are comparable to NPN. However, for the \textit{Ionosphere} data, NPN has much better results in terms of both the error rates and MCC. 

\section{Multi-category Classification: Simulation Study} \label{sec:multsim}

\begin{table}[t]
\centering
\caption{Classification error rates for logistic transformation. }
\resizebox{0.70\textwidth}{!}{
\begin{tabular}{@{}llllllll@{}}
\toprule
\textbf{p} & $\mathbf{n_L}$ & \textbf{K} & \textbf{n} & \textbf{NPN} & \textbf{RF} & \textbf{KNN} & \textbf{ROCC} \\ \midrule
10        & 5  & 5  & 40-60  & \textbf{0.2733} & 0.4156 & 0.4382 & 0.7271 \\
\textbf{} &    &    &        & ($\pm$0.2341) & ($\pm$0.0563) & ($\pm$0.0818) & ($\pm$0.039)  \\ \hline
10        & 5  & 10 & 40-60  & \textbf{0.1649} & 0.5306 & 0.5352 & 0.8701 \\
\textbf{} &    &    &        & ($\pm$0.0921) & ($\pm$0.0356) & ($\pm$0.0406) & ($\pm$0.0202) \\ \hline
15        & 10 & 5  & 80-120 & 0.5083 & \textbf{0.3312} & 0.3353 & 0.7204 \\
\textbf{} &    &    &        & ($\pm$0.2758) & ($\pm$0.0427) & ($\pm$0.0431) & ($\pm$0.0248) \\ \hline
15        & 10 & 10 & 80-120 & \textbf{0.2793} & 0.4506 & 0.4584 & 0.879  \\
          &    &    &        & ($\pm$0.1154) & ($\pm$0.0347) & ($\pm$0.0421) & ($\pm$0.0133) \\ \hline
30        & 15 & 5  & 80-120 & 0.3248 & 0.3401 & \textbf{0.303}  & 0.7177 \\
          &    &    &        & ($\pm$0.3141) & ($\pm$0.0491) & ($\pm$0.0307) & ($\pm$0.0228) \\ \hline
30        & 15 & 10 & 80-120 & \textbf{0.0681}      & 0.4593 & 0.4355 & 0.8708 \\
          &    &    &        & ($\pm$0.1444)      & ($\pm$0.0286) & ($\pm$0.0199) & ($\pm$0.0125) \\ \hline
15        & 15 & 5  & 80-120 & 0.4037 & 0.2814 & \textbf{0.2714} & 0.7176 \\
          &    &    &        & ($\pm$0.2456) & ($\pm$0.0502) & ($\pm$0.0468) & ($\pm$0.0338) \\ \hline
15        & 15 & 10 & 80-120 & \textbf{0.0622} & 0.4049 & 0.3935 & 0.8735 \\
          &    &    &        & ($\pm$0.0825) & ($\pm$0.0287) & ($\pm$0.0302) & ($\pm$0.0089) \\ \bottomrule
\end{tabular}
}
\label{tab:multl}
\end{table}

\begin{table}[!b]
\centering
\caption{Classification error rates for normal transformation. }
\resizebox{0.70\textwidth}{!}{
\begin{tabular}{@{}llllllll@{}}
\toprule
\textbf{p} & $\mathbf{n_L}$ & \textbf{K} & \textbf{n} & \textbf{NPN} & \textbf{RF} & \textbf{KNN} & \textbf{ROCC} \\ \midrule
10        & 5  & 5  & 40-60  & \textbf{0.2927} & 0.4157 & 0.4229 & 0.7321 \\ 
\textbf{} &    &    &        & ($\pm$0.2134) & ($\pm$0.0604) & ($\pm$0.0663) & ($\pm$0.0394) \\ \hline
10        & 5  & 10 & 40-60  & \textbf{0.1835} & 0.5239 & 0.5373 & 0.8711 \\
\textbf{} &    &    &        & ($\pm$0.1119) & ($\pm$0.0298) & ($\pm$0.0477) & ($\pm$0.0208) \\ \hline
15        & 10 & 5  & 80-120 & 0.4758 & \textbf{0.3357} & 0.3358 & 0.7183 \\
\textbf{} &    &    &        & ($\pm$0.204)  & ($\pm$0.0432) & ($\pm$0.0473) & ($\pm$0.0237) \\ \hline
15        & 10 & 10 & 80-120 & \textbf{0.2824} & 0.4512 & 0.4593 & 0.8787 \\
          &    &    &        & ($\pm$0.158)  & ($\pm$0.0373) & ($\pm$0.0383) & ($\pm$0.0108) \\ \hline
30        & 15 & 5  & 80-120 & 0.7395 & 0.332  & \textbf{0.2972} & 0.7138 \\
          &    &    &        & ($\pm$0.1089) & ($\pm$0.0365) & ($\pm$0.0303) & ($\pm$0.0258) \\ \hline
30        & 15 & 10 & 80-120 & 0.8565      & 0.4585 & \textbf{0.4296} & 0.8678 \\
          &    &    &        & ($\pm$0.0525)  & ($\pm$0.0344) & ($\pm$0.0226) & ($\pm$0.0132) \\ \hline
15        & 15 & 5  & 80-120 & 0.3035 & 0.2875 & \textbf{0.2804} & 0.7242 \\
          &    &    &        & ($\pm$0.2336) & ($\pm$0.0593) & ($\pm$0.0482) & ($\pm$0.0378) \\ \hline
15        & 15 & 10 & 80-120 & \textbf{0.0844} & 0.4074 & 0.3999 & 0.8752 \\
          &    &    &        & ($\pm$0.0926) & ($\pm$0.0347) & ($\pm$0.0326) & ($\pm$0.0117) \\ \bottomrule
\end{tabular}
}
\label{tab:multn}
\end{table}

Given $K$ (say) classes and a given dimension $p$, we simulate the observations from $K$ multivariate normal distributions $\N_p(\boldsymbol{\mu}_k, \boldsymbol{\Sigma}_k)$ where $k=1,\cdots,K$. Like the simulation setup in Section \ref{sec:binsim}, means and covariances of the underlying Gaussian distributions are randomly generated as follows:
\begin{itemize}
    \item Generate $\mu_{kd}$ independently and identically from a uniform distribution ranging in $[0,4]$ for $d=1, \ldots, p$; $k=1, \ldots, K$.
    \item Generate $\boldsymbol{\Sigma}_k$ by $\boldsymbol{A}^{\mathrm{T}}\boldsymbol{A}$, where every element in $\boldsymbol{A}$ is generated from the uniform distribution over $[-1,1]$; $k=1, \ldots, K$.
\end{itemize}

We considered two true transformations: 
\begin{enumerate}
    \item \textit{The Logistic transformation}: $\left\{1+\exp\left(-\frac{X_d-\left(\sum_{k=1}^K \mu_{kd}\right)/K}{\left(\sum_{k=1}^K \sqrt{\Sigma_{k d,d}}\right)/K}\right)\right\}^{-1}$ for $d=1, \ldots, p$;
    \item \textit{The Normal transformation}: $\left(\left(\sum_{k=1}^K \sqrt{\Sigma_{k d,d}}\right)/K\right)^{-1}\Phi\left(\frac{X_d-\left(\sum_{k=1}^K \mu_{kd}\right)/K}{\left(\sum_{k=1}^K \sqrt{\Sigma_{k d,d}}\right)/K}\right)$ for $d=1, \ldots, p$. 
\end{enumerate}

For each of the settings above, we take $p$ ranging between $5$ and $30$ and generate $n$ number of observations for each class where $n$ is between $40$--$60$ or $80$--$120$. We select $n_L$ as $5$ or $15$, where $n_L$ is the number of labeled data points in each class. We generate $1000$ data points for each class to test the model. Each experiment is repeated $15$ times. We note the classification error rates and report their mean and standard error (within parenthesis) in Tables \ref{tab:multl} and \ref{tab:multn}. We compare our method with some of the best supervised learning methods and find that our method produces competitive results and is particularly useful when the total number of observations and labeled observations are very few. The methods used for comparison purposes are Random Forest (RF), $K$-Nearest Neighbours (KNN) and ROC-AUC classifier (ROCC). We use the R package \texttt{caret} (\citet{Kuhn2008}) to fit these models.

\section{Discussion} \label{sec:diss}
Our proposed method gives good competitive results in both simulation studies and real data analysis when compared with some of the best semi-supervised machine learning methods in a binary classification setup. Further, the results obtained from our method is also competitive in a multi-class classification setup. The usage of conjugate normal priors and data augmentation technique in the Gibbs sampling algorithm makes the proposed method quite easily computable. The assumptions on the data are minimal making the proposed method workable on any semi-supervised classification dataset without much worry. In conclusion, this method is a safe choice for semi-supervised binary classification problems. 


\bibliographystyle{plainnat}
\bibliography{references}

\begin{thebibliography}{23}
\providecommand{\natexlab}[1]{#1}
\providecommand{\url}[1]{\texttt{#1}}
\expandafter\ifx\csname urlstyle\endcsname\relax
  \providecommand{\doi}[1]{doi: #1}\else
  \providecommand{\doi}{doi: \begingroup \urlstyle{rm}\Url}\fi

\bibitem[Balcan et~al.(2005)Balcan, Blum, Choi, Lafferty, Pantano, Rwebangira, and Zhu]{Balcan2005}
M.-F. Balcan, A.~Blum, P.~Choi, J.D. Lafferty, B.~Pantano, M.R. Rwebangira, and X.~Zhu.
\newblock Person identification in webcam images: An application of semi-supervised learning.
\newblock In \emph{Intl Workshop Learning with Partially Classified Training Data}, 2005.

\bibitem[Belkin et~al.(2006)Belkin, Niyogi, and Sindhwani]{Belkin2006}
M.~Belkin, P.~Niyogi, and V.~Sindhwani.
\newblock Manifold regularization: A geometric framework for learning from labeled and unlabeled examples.
\newblock \emph{Journal of Machine Learning Research}, 7:\penalty0 2399--2434, 2006.

\bibitem[Blum and Mitchell(1998)]{Blum1998}
A.~Blum and T.~Mitchell.
\newblock Combining labeled and unlabeled data with co-training.
\newblock In \emph{Proceedings of the Eleventh Annual Conference on Computational Learning Theory}, pages 92--100, 1998.

\bibitem[Cozman et~al.(2003)Cozman, Cohen, and Cirelo]{Cozman2003}
F.G. Cozman, I.~Cohen, and M.C. Cirelo.
\newblock Semi-supervised learning of mixture models.
\newblock In \emph{Proceedings of the 20th International Conference on Machine Learning (ICML-03)}, pages 99--106, 2003.

\bibitem[Dempster et~al.(1977)Dempster, Laird, and Rubin]{Dempster1977}
A.P. Dempster, N.M. Laird, and D.B. Rubin.
\newblock Maximum likelihood from incomplete data via the em algorithm.
\newblock \emph{Journal of the Royal Statistical Society. Series B (Methodological)}, 39\penalty0 (1):\penalty0 1--22, 1977.

\bibitem[Grandvalet and Bengio(2005)]{Grandvalet2005}
Y.~Grandvalet and Y.~Bengio.
\newblock Semi-supervised learning by entropy minimization.
\newblock In \emph{Advances in Neural Information Processing Systems}, pages 529--536, 2005.

\bibitem[Hein and Maier(2007)]{Hein2007}
M.~Hein and M.~Maier.
\newblock Manifold denoising.
\newblock In \emph{Advances in Neural Information Processing Systems}, pages 561--568, 2007.

\bibitem[Krijthe(2017)]{Krijthe2017}
Jesse~H Krijthe.
\newblock Rssl: Semi-supervised learning in r.
\newblock In \emph{Reproducible Research in Pattern Recognition: First International Workshop, RRPR 2016, Canc{\'u}n, Mexico, December 4, 2016, Revised Selected Papers 1}, pages 104--115. Springer, 2017.

\bibitem[Krijthe(2014)]{Krijthe2014}
J.H. Krijthe.
\newblock Rssl: Semi-supervised learning in r.
\newblock In \emph{International Workshop on Reproducible Research in Pattern Recognition}, pages 104--115, 2014.

\bibitem[Krijthe and Loog(2015)]{Krijthe2015}
J.H. Krijthe and M.~Loog.
\newblock Implicitly constrained semi-supervised least squares classification.
\newblock In \emph{International Symposium on Intelligent Data Analysis}, pages 158--169, 2015.

\bibitem[Kuhn(2008)]{Kuhn2008}
Max Kuhn.
\newblock Building predictive models in r using the caret package.
\newblock \emph{Journal of Statistical Software}, 28\penalty0 (5):\penalty0 1–26, 2008.
\newblock \doi{10.18637/jss.v028.i05}.

\bibitem[Lawrence and Platt(2005)]{Lawrence2005}
N.D. Lawrence and J.C. Platt.
\newblock Learning through model combination.
\newblock In \emph{Proceedings of the International Conference on Artificial Intelligence and Statistics (AISTATS)}, pages 312--319, 2005.

\bibitem[Li and Ghosh(2015)]{Li2015}
Yifang Li and Sujit~K Ghosh.
\newblock Efficient sampling methods for truncated multivariate normal and student-t distributions subject to linear inequality constraints.
\newblock \emph{Journal of Statistical Theory and Practice}, 9:\penalty0 712--732, 2015.

\bibitem[Li et~al.(2013)Li, Tsang, Kwok, and Zhou]{Li2013}
Yu-Feng Li, Ivor~W Tsang, James~T Kwok, and Zhi-Hua Zhou.
\newblock Convex and scalable weakly labeled svms.
\newblock \emph{Journal of Machine Learning Research}, 14\penalty0 (7), 2013.

\bibitem[Liu et~al.(2012)Liu, Han, Yuan, Lafferty, and Wasserman]{Liu2012}
H.~Liu, F.~Han, M.~Yuan, J.~Lafferty, and L.~Wasserman.
\newblock High-dimensional semiparametric gaussian copula graphical models.
\newblock \emph{The Annals of Statistics}, 40\penalty0 (4):\penalty0 2293--2326, 2012.

\bibitem[Loog(2010)]{Loog2010}
M.~Loog.
\newblock Constrained parameter estimation for semi-supervised learning: the case of the nearest mean classifier.
\newblock In \emph{Joint European Conference on Machine Learning and Knowledge Discovery in Databases}, pages 291--304, 2010.

\bibitem[Loog(2011)]{Loog2011}
M.~Loog.
\newblock Semi-supervised linear discriminant analysis using moment constraints.
\newblock In \emph{Iapr International Workshop on Partially Supervised Learning}, pages 32--41, 2011.

\bibitem[Mulgrave and Ghosal(2020)]{Mulgrave2020}
J.~Mulgrave and S.~Ghosal.
\newblock Bayesian inference for high-dimensional data.
\newblock \emph{Bayesian Analysis}, 15:\penalty0 445--473, 2020.

\bibitem[Sindhwani and Keerthi(2006)]{Sindhwani2006}
V.~Sindhwani and S.S. Keerthi.
\newblock Large scale semi-supervised linear svms.
\newblock In \emph{Proceedings of the 29th Annual International ACM Sigir Conference on Research and Development in Information Retrieval}, pages 477--484, 2006.

\bibitem[Yarowsky(1995)]{Yarowsky1995}
D.~Yarowsky.
\newblock Unsupervised word sense disambiguation rivaling supervised methods.
\newblock In \emph{33rd Annual Meeting of the Association for Computational Linguistics}, pages 189--196, 1995.

\bibitem[Zemel and Carreira-Perpin{\'a}n(2005)]{Zemel2005}
R.S. Zemel and M.{\'A}. Carreira-Perpin{\'a}n.
\newblock Proximity graphs for clustering and manifold learning.
\newblock In \emph{Advances in Neural Information Processing Systems}, pages 225--232, 2005.

\bibitem[Zhang and Lee(2007)]{Zhang2007}
X.~Zhang and W.S. Lee.
\newblock Hyperparameter learning for graph based semi-supervised learning algorithms.
\newblock In \emph{Advances in Neural Information Processing Systems}, pages 1585--1592, 2007.

\bibitem[Zhu(2005)]{Zhu2005}
X.~Zhu.
\newblock Semi-supervised learning literature survey.
\newblock Technical Report Tech. Rep. No. 1530, Computer Sciences, University of Wisconsin-Madison, 2005.

\end{thebibliography}
\end{document}